\documentclass[10pt,twocolumn,letterpaper]{article}
\usepackage[svgnames,table]{xcolor}
\usepackage{cvpr}
\usepackage{times}
\usepackage{epsfig}
\usepackage{graphicx}
\usepackage{gensymb} 
\usepackage{color}
\usepackage[pagebackref=true,breaklinks=true,letterpaper=true,colorlinks,bookmarks=false]{hyperref}
\usepackage{amsmath,amsfonts,amssymb,mathrsfs}
\usepackage{makecell}

\usepackage{graphicx}
\usepackage[ruled]{algorithm2e}
\usepackage{amsmath,amssymb}
\usepackage{epstopdf}
\usepackage{tabularx}
\usepackage{eqnarray}
\usepackage{array,booktabs}
\usepackage{enumitem}
\usepackage[space]{grffile} 
\usepackage{soul} 
\usepackage{float}
\usepackage{tabularx,ragged2e} 
\usepackage{bbm}
\usepackage{pifont} 
\usepackage{cancel}
\usepackage{wrapfig}
\usepackage{color, colortbl}
\usepackage{tikz} 

\makeatletter

\renewcommand{\paragraph}{%
  \@startsection{paragraph}{4}%
  {\z@}{0.3ex \@plus 1ex \@minus .2ex}{-1em}
  {\normalfont\normalsize\bfseries}%
}
\makeatother

\definecolor{myGray}{rgb}{0.6,0.6,0.6}
\definecolor{myBlue}{rgb}{0.1,0.1,0.8}

\newcommand{\RR}{\mathbb{R}}

\DeclareMathOperator*{\argmax}{arg\,max}
\DeclareMathOperator*{\softmax}{softmax}

\def\onedot{\ifx\@let@token.\else.\null\fi\xspace}

\def\eg{\emph{e.g}\onedot} 
\def\ie{\emph{i.e}\onedot}

\def\etal{\emph{et al}\onedot}

\definecolor{pcGray}{rgb}{0.5,0.5,0.5}

\definecolor{pccGray}{rgb}{0.3,0.3,0.3}

\newcommand{\fig}{Fig.~}
\newcommand{\eq}{Eq.\,}
\newcommand{\sect}{Section~}

\usepackage{listings}
\usepackage{color}

\ifx \@etal \@empty \newcommand{\etal}{et al.} \fi
\ifx \@eg \@empty \newcommand{\eg}{e.g.,~} \fi
\ifx \@ie \@empty \newcommand{\ie}{i.e.,~} \fi
\ifx \@etc \@empty  \fi

\newcommand{\M}{{\cal M}}

\newcommand{\bb}{\boldsymbol{b}}

\newcommand{\bh}{{\boldsymbol{h}}}

\newcommand{\bq}{{\boldsymbol{q}}}

\newcommand{\bs}{{\boldsymbol{s}}}

\newcommand{\bv}{{\boldsymbol{v}}}
\newcommand{\bw}{{\boldsymbol{w}}}

\def\thmcolon{\hspace{-.85em} {\bf :}}

\newtheorem{THEOREM}{Theorem}[section]
\newtheorem{LEMMA}[THEOREM]{Lemma}
\newtheorem{PROPOSITION}[THEOREM]{Proposition}
\newtheorem{COROLLARY}[THEOREM]{Corollary}
\newtheorem{DEFINITION}[THEOREM]{Definition}
\newtheorem{OBSERVATION}[THEOREM]{Observation}

\definecolor{bgCode}{rgb}{0.94, 0.94, 1.0}
\graphicspath{{figures/},{plots/}}

\cvprfinalcopy
\def\cvprPaperID{20}

\ifcvprfinal\fi

\newcommand{\vvspace}[1]{\vspace{#1}}

\newcommand{\setfont}[1]{{\mathcal{#1}}} 
\newcommand{\loss}{\mathscr{L}} 
\renewcommand{\M}{\setfont{M}} 
\newcommand{\ProtoSet}{\mathit{\Phi}}  
\newcommand{\proto}{\boldsymbol{\phi}}
\newcommand{\transParams}{{\boldsymbol{\theta}}}

\makeatletter
\let\@fnsymbol\@arabic
\makeatother

\begin{document}

\title{Visual Question Answering as a Meta Learning Task}

\author{Damien Teney~~~~~~Anton van den Hengel\\
Australian Centre for Visual Technologies\\
The University of Adelaide\\
{\tt\small \{damien.teney,anton.vandenhengel\}@adelaide.edu.au}
}

\maketitle

\begin{abstract}
The predominant approach to Visual Question Answering (VQA) demands that the model represents within its weights all of the information required to answer any question about any image. Learning this information from any real training set seems unlikely, and representing it in a reasonable number of weights doubly so. We propose instead to approach VQA as a meta learning task, thus separating the question answering method from the information required. At test time, the method is provided with a support set of example questions/answers, over which it reasons to resolve the given question. The support set is not fixed and can be extended without retraining, thereby expanding the capabilities of the model. To exploit this dynamically provided information, we adapt a state-of-the-art VQA model with two techniques from the recent meta learning literature, namely prototypical networks and  meta networks. Experiments demonstrate the capability of the system to learn to produce completely novel answers (\ie never seen during training) from examples provided at test time. 
In comparison to the existing state of the art, the proposed method produces qualitatively distinct results with higher recall of rare answers, and a better sample efficiency that allows training with little initial data. More importantly, it represents an important step towards vision-and-language methods that can learn and reason on-the-fly.
\end{abstract}


\section{Introduction}
\label{sec:intro}

The task of Visual Question Answering (VQA) demands that an agent correctly answer a previously unseen question about a previously unseen image. The fact that neither the question nor the image is specified until test time means that the agent must embody most of the achievements of Computer Vision and Natural Language Processing, and many of those of Artificial Intelligence.

VQA is typically framed in a purely supervised learning setting. A large training set of example questions, images, and their correct answers is used to train a method to map a question and image to scores over a predetermined, fixed vocabulary of possible answers using the maximum likelihood~\cite{wu2017survey}. This approach has inherent scalability issues, as it attempts to represent all world knowledge within the finite set of parameters of a model such as deep neural network. Consequently, a trained VQA system can only be expected to produce correct answers to questions from a very similar distribution to those in the training set. Extending the model knowledge or expanding its domain coverage is only possible by retraining it from scratch, which is computationally costly, at best. This approach is thus fundamentally incapable of fulfilling the ultimate promise of VQA, which is answering general questions about general images.

\begin{figure}[t]
  \centering
  \includegraphics[width=0.94\linewidth]{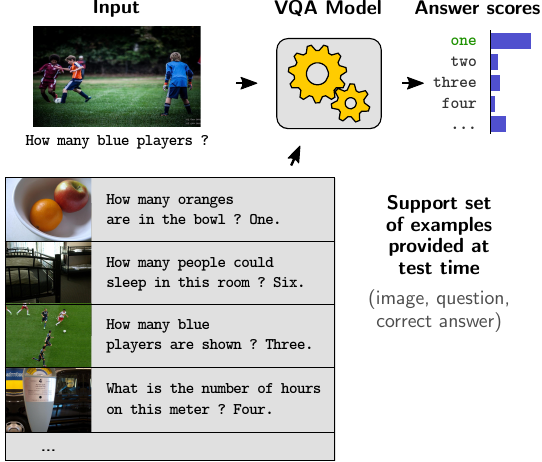}
  \label{fig:teaser}
  \vvspace{7pt}
  \caption{This paper considers Visual Question Answering in a meta learning setting. The model is initially trained on a small set of questions/answers, and is provided with a, possibly large, additional \emph{support set} of examples at test time. The model must \emph{learn to learn}, or to exploit the additional data on-the-fly, without the need for retraining the model. Notably, performance improves as additional and more relevant examples are included.}
  \vvspace{-8pt}
\end{figure}

As a solution to these issues we propose a meta-learning approach to the problem. 
The meta learning approach implies that the model \emph{learns to learn}, \ie it learns to use a set of examples provided at test time to answer the given question (\fig\ref{fig:teaser}). Those examples are questions and images, each with their correct answer, such as might form part of the training set in a traditional setting. They are referred to here as the \emph{support set}. Importantly, the support set is not fixed.  Note also that the support set may be large, and that the majority of its elements may have no relevance to the current question.  It is provided to the model at test time, and can be expanded with additional examples to increase the capabilities of the model.  
The model we propose `learns to learn' in that it is able to identify and exploit the relevant examples within a potentially large support set dynamically, at test time.
Providing the model with more information thus does not require retraining, and the ability to exploit such a support set greatly improves the practicality and scalability of the system. Indeed, it is ultimately desirable for a practical VQA system to be adaptable to new domains and to continuously improve as more data becomes available. That vision is a long term objective and this work takes only a small step in that direction.

Our primary technical contribution is to adapt a state-of-the-art VQA model\cite{teney2017challenge} to the meta learning scenario. Our resulting model is a deep neural network that uses sets of dynamic parameters --~also known as fast weights~-- determined at test time depending on the provided support set. The dynamic parameters allow to modify adaptively the computations performed by the network and adapt its behaviour depending on the support set. We perform a detailed study to evaluate the effectiveness of those techniques under various regimes of training and support set sizes. Those experiments are based on the VQA v2 benchmark, for which we propose data splits appropriate to study a meta learning setting.

A completely new  capability demonstrated by the resulting system is to learn to produce completely novel answers (\ie answers not seen during training). Those new answers are only demonstrated by instances of the support set provided at test time. In addition to these new capabilities, the system exhibits a qualitatively distinct behaviour to existing VQA systems in its improved handling of rare answers. Since datasets for VQA exhibit a heavy class imbalance, with a small number of answers being much more frequent than most others, models optimized for current benchmarks are prone to fall back on frequent ``safe'' answers. In contrast, the proposed model is inherently less likely to fall victim to dataset biases, and exhibits a higher recall over rare answers. The proposed model does \emph{not} surpass existing methods on the common aggregate accuracy metric, as is to be expected given that it does not overfit to the dataset bias, but it nonetheless exhibits desirable traits overall.

\noindent
The contributions of this paper are summarized as follows.
\setlist{nolistsep,leftmargin=*}
\begin{enumerate}[noitemsep]
  \item We re-frame VQA as a meta learning task, in which the model is provided a test time with a support set of supervised examples (questions and images with their correct answers).

  \item We describe a neural network architecture and training procedure able to leverage the meta learning scenario. The model is based on a state-of-the-art VQA system and takes inspiration in techniques from the recent meta learning literature, namely prototypical networks~\cite{snell2017proto} and meta networks~\cite{munkhdalaiY17meta}.

  \item We provide an experimental evaluation of the proposed model in different regimes of training and support set sizes and across variations in design choices.

  \item Our results demonstrate the unique capability of the model to produce novel answers, \ie answers never seen during training, by learning from support instances, an improved recall of rare answers, and a better sample efficiency than existing models.

\end{enumerate}

\section{Related Work}
\label{sec:related}

\paragraph{Visual question answering}
Visual question answering has gathered significant interest from the computer vision community~\cite{antol2015vqa}, as it constitutes a practical setting to evaluate deep visual understanding. In addition to visual parsing, VQA requires the comprehension of a text question, and combined reasoning over vision and language, sometimes on the basis of external or common-sense knowledge. See~\cite{wu2017survey} for a recent survey of methods and datasets.

VQA is always approached in a supervised setting, using large datasets~\cite{antol2015vqa,goyal2016balanced,krishnavisualgenome,zhu2015visual7w} of human-proposed questions with their correct answers to train a machine learning model. The \emph{VQA-real} and \emph{VQA v2} datasets~\cite{antol2015vqa,goyal2016balanced} have served as popular benchmarks by which to evaluate and compare methods. Despite the large scale of those datasets, \eg more than 650,000 questions in \emph{VQA v2}, several limitations have been recognized. These relate to the dataset bias (\ie the non-uniform, long-tailed distribution of answers) and the question-conditioned bias (making answers easy to guess given a question without the image). For example, the answer \emph{Yes} is particularly prominent in~\cite{antol2015vqa} compared to \emph{no}, and questions starting with \emph{How many} can be answered correctly with the answer \emph{two} more than 30\% of the time~\cite{goyal2016balanced}. These issues plague development in the field by encouraging methods which fare well on common questions and concepts, rather than on rare answers or more complicated questions. The aggregate accuracy metric used to compare methods is thus a poor indication of method capabilities for visual understanding. Improvements to datasets have been introduced~\cite{agrawal2017c,goyal2016balanced,zhang2015balanced}, including the \emph{VQA v2}, but they only partially solve the evaluation problems. An increased interest has appeared in the handling of rare words and answers~\cite{rama2017vqa,teney2016zsvqa}. The model proposed in this paper is inherently less prone to incorporate dataset biases than existing methods, and shows superior performance for handling rare answers. It accomplishes this by keeping a memory made up of explicit representations of training and support instances.

\paragraph{VQA with additional data}
In the classical supervised setting, a fixed set of questions and answers is used to train a model once and for all. With few exceptions, the performance of such a model is fixed as it cannot use additional information at test time. Among those exceptions,~\cite{wu2015ask,wang2015explicit} use an external knowledge base to gather non-visual information related to the input question. In~\cite{teney2016zsvqa}, the authors use visual information from web searches in the form of exemplar images of question words, and better handle rare and novel words appearing in questions as a result. In~\cite{teney2017challenge}, the same authors use similar images from web searches to obtain visual representations of candidate answers. 



Those methods use ad-hoc engineered techniques to incorporate external knowledge in the VQA model. In comparison, this paper presents a much more general approach. We expands the model knowledge with data provided in the form of additional supervised examples (questions and images with their correct answer). A demonstration of the broader generality of our framework over the works above is its ability to produce novel answers, \ie never observed during initial training and learned only from test-time examples.

Recent works on text-based question answering have investigated the retrieval of external information with reinforcement learning~\cite{nogueira2017task,narasimhan2016improving,buck2017ask}. Those works are tangentially related and complementary to the approach explored in this paper.

\paragraph{Meta learning and few shot learning}
The term \emph{meta learning} broadly refers to methods that \emph{learn to learn}, \ie that train models to make better use of training data. It applies to approaches including the learning of gradient descent-like algorithms such as~\cite{andrychowicz2016learning,finn2017model,hochreiter2001learning,ravi2016optimization} for faster training or fine-tuning of neural networks, and the learning of models that can be directly fed training examples at test time~\cite{BertinettoHVTV16,snell2017proto,triantafillou2017few}. The method we propose falls into the latter category. Most works on meta learning are motivated by the challenge of one-shot and few-shot visual recognition, where the task is to classify an image into categories defined by a few examples each. Our meta learning setting for VQA bears many similarities. VQA is treated as a classification task, and we are provided, at test time, with examples that illustrate the possible answers~--~possibly a small number per answer. Most existing methods are, however, not directly applicable to our setting, due to the large number of classes (\ie possible answers), the heavy class imbalance, and the need to integrate into an architecture suitable to VQA. For example, recent works such as~\cite{triantafillou2017few} propose efficient training procedures that are only suitable for a small number of classes.


Our model uses a set of memories within a neural network to store the activations computed over the support set. Similarly, Kaiser \etal~\cite{kaiser2017rare} store past activations to remember ``rare events'', which was notably evaluated on machine translation. Our model also uses network layers parametrized by dynamic weights, also known as fast weights. Those are determined at test time depending on the actual input to the network. Dynamic parameters have a long history in neural networks~\cite{schmidhuber2008learning} and have been used previously for few-shot recognition~\cite{BertinettoHVTV16} and for VQA~\cite{Noh2015image}. One of the memories within our network stores the gradient of the loss with respect to static weights of the network, which is similar to the Meta Networks model proposed by Munkhdalai \etal~\cite{munkhdalaiY17meta}. Finally, our output stage produces scores over possible answers by similarity to prototypes representing the output classes (answers). This follows a similar idea to the Prototypical Networks~\cite{snell2017proto}.

\paragraph{Continuum learning}
An important outcome of framing VQA in a meta learning setting is to develop models capable of improving as more data becomes available. This touches the fields of incremental~\cite{fernando17pathnet,rebuffi2016icarl} and continuum learning~\cite{aljundi2016expert,ranzato2016continuum,yoon2017lifelong}. Those works focus on the fine-tuning of a network with new training data, output classes and/or tasks. In comparison, our model does not modify itself over time and cannot experience negative domain shift or catastrophic forgetting, which are a central concern of continuum learning~\cite{kirk2016catastrophic}. Our approach is rather to use such additional data on-the-fly, at test time, \ie without an iterative retraining. An important motivation for our framework is its potential to apply to support data of a different nature than question/answer examples. We consider this to be an important direction for future work. This would allow to leverage general, non VQA-specific data, \eg from knowledge bases or web searches.


\section{VQA in a Meta Learning Setting}

\begin{figure*}[t]
  \centering
  \includegraphics[width=0.93\textwidth]{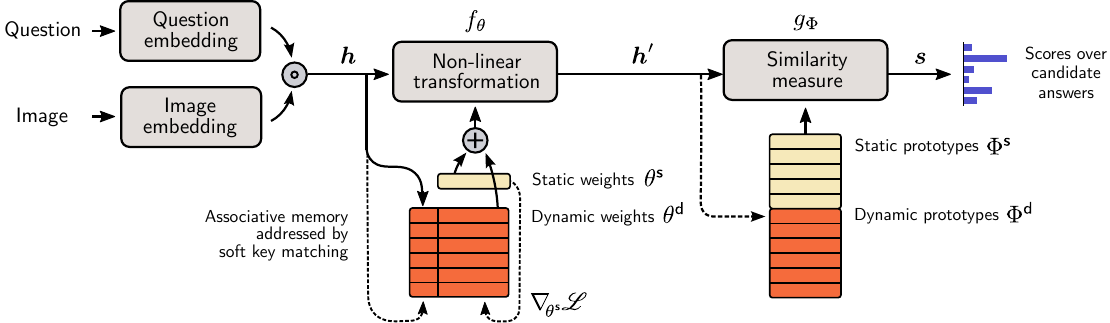}
  \label{fig:network}
  \vvspace{5pt}
  \caption{Overview of the proposed model. We obtain an embedding the input question and image following~\cite{teney2017challenge} and our contributions concern the mapping of this embedding to scores over a set of candidate answers. First, a \textbf{non-linear transformation} (implemented as a gated hyperbolic tangent layer) is parametrized by static and dynamic weights. Static ones are learned like traditional weights by gradient descent, while dynamic ones are determined based on the actual input and a memory of candidate dynamic weights filled by processing the support set. Second, a \textbf{similarity measure} compares the resulting feature vector to a set of prototypes, each representing a specific candidate answer. Static prototypes are learned like traditional weights, while dynamic prototypes are determined by processing the support set. Dashed lines indicate data flow during the processing of the support set. See \sect\ref{sec:network} for details.}
  \vvspace{-4pt}
\end{figure*}

The traditional approach to VQA is in a supervised setting described as follows. A model is trained to map an input question $\mathsf{Q}$ and image $\mathsf{I}$ to scores over candidate answers~\cite{wu2017survey}. The model is trained to maximize the likelihood of correct answers over a training set $\setfont{T}$ of triplets $(\mathsf{Q},\mathsf{I},\hat{\bs})$, where $\boldsymbol{\hat{s}}\in [0,1]^A$ represents the vector of ground truth scores of the predefined set of $A$ possible answers. At test time, the model is evaluated on another triplet $(\mathsf{Q}',\mathsf{I}',\hat{\bs}')$ from an evaluation or test set $\setfont{E}$. The model predicts scores $\bs'$ over the set of candidate answers, which can be compared to the ground truth $\hat{\bs}'$ for evaluation purposes.

We extend the formulation above to a meta learning setting by introducing an additional \emph{support set} $\setfont{S}$ of similar triplets $(\mathsf{Q}'',\mathsf{I}'',\hat{\bs}'')$. These are provided to the model at test time. At a minimum, we define the support set to include the training examples themselves, \ie $\setfont{S}=\setfont{T}$, but more interestingly, the support set can include novel examples $\setfont{S'}$ provided at test time. They constitute additional data to learn from, such that $\setfont{S}=\setfont{T}\cup\setfont{S'}$. The triplets $(\mathsf{Q},\mathsf{I},\hat{\bs})$ in the support set can also include novel answers, never seen in the training set. In that case, the ground truth score vectors $\hat{bs}$ of the other elements in the support are simply padded with zeros to match the larger size $A'$ of the extended set of answers.

The following sections describe a deep neural network that can take advantage of the support set at test time. To leverage the information contained in the support set, the model must learn to utilize these examples on-the-fly at test time, without retraining of the whole model.

\section{Proposed Model}
\label{sec:network}

The proposed model (\fig\ref{fig:network}) is a deep neural network that extends the state-of-art VQA system of Teney \etal\cite{teney2017challenge}. Their system implements the joint embedding approach common to most modern VQA models~\cite{wu2017survey,yang2015stacked,jabri2016revisiting,kazemi2017baseline}, which is followed by a multi-label classifier over candidate answers. Conceptually, we separate the architecture into (1)~the embedding part that encodes the input question and image, and (2)~the classifier part that handles the reasoning and actual question answering\footnote{The separation of the network into an embedding and a classifier parts is conceptual. The division is arbitrarily placed after the fusion of the question and image embeddings. Computational requirements aside, the concept of dynamic parameters is in principle applicable to earlier layers as in~\cite{BertinettoHVTV16}.}. The contributions of this paper address only the second part. Our contributions are orthogonal to developments on the embedding part, which could also benefit \eg from advanced attention mechanisms or other computer vision techniques~\cite{anderson2017features,wang2016machine,wu2017survey}. We follow the implementation of~\cite{teney2017challenge} for the embedding part. For concreteness, let us mention that the question embedding uses \textit{GloVe} word vectors~\cite{pennington2014glove} and a Recurrent Gated Unit (GRU~\cite{cho2014learning}). The image embedding uses features from a CNN (Convolutional Neural Network) with bottom-up attention~\cite{anderson2017features} and question-guided attention over those features. See~\cite{teney2017challenge} for details.

For the remainder of this paper, we abstract the embedding to modules that produce respectively the question and image vectors $\bq$ and $\bv \in \RR^D$. They are combined with a Hadamard (element-wise) product into $\bh = \bq ~\circ~ \bv$, which forms the input to the classifier on which we now focus on. The role of the classifier is to map $\bh$ to a vector of scores $\bs\in [0,1]^A$ over the candidate answers. We propose a definition of the classifier that generalizes the implementation of traditional models such as~\cite{teney2017challenge}. The input to the classifier $\bh\in \RR^D$ is first passed through a non-linear transformation $f_\transParams: \RR^D\to \RR^D$, then through a mapping to scores over the set of candidate answers $g_\ProtoSet: \RR^D\to [0,1]^A$. This produces a vector of predicted scores $\bs = f_\transParams(g_\ProtoSet(\bh))$. In traditional models, the two functions correspond to a stack of non-linear layers for $f_\transParams$, and a linear layer followed by a softmax or sigmoid for $g_\ProtoSet$. We now show how to extend $f_\transParams$ and $g_\ProtoSet$ to take advantage of the meta learning setting.

\subsection{Non-linear Transformation $f_\transParams(\cdot)$}

The role of the non-linear transformation $f_\transParams(\bh)$ is to map the embedding of the question/image $\bh$ to a representation suitable for the following (typically linear) classifier. This transformation can be implemented in a neural network with any type of non-linear layers. Our contributions are agnostic to this implementation choice. We follow~\cite{teney2017challenge} and use a gated hyperbolic tangent layer~\cite{dauphin2016languagecnn}, defined as
\abovedisplayskip=4pt
\belowdisplayskip=4pt
\begin{align}
  \label{eq:nonLinearLayer}
  f_\transParams(\bh) ~~=~~ \sigma( W \bh + \bb ) ~\circ~ \tanh\,(W' \bh + \bb' )
\end{align}
where $\sigma$ is the logistic activation function, $W, W' \in \RR^{D \times D}$ are learned weights, $\bb, \bb' \in \RR^D$ are learned biases, and~$\circ$~is the Hadamard (element-wise) product. For notation purposes, we define the parameters $\transParams$ as the concatenation of the vectorized weights and biases, \ie $\transParams=[W_:;W'_:;\bb;\bb']$. This vector thus contains all of the weights and biases used by the non-linear transformation. A traditional model would learn the weights $\transParams$ by backpropagation and gradient descent on the training set, and they would be held static during test time. We propose instead to adaptively adjust the weights at test time, depending on the input $\bh$ and  the available support set. Concretely, we use a combination of static parameters $\transParams^\mathsf{s}$ learned in the traditional manner, and dynamic ones $\transParams^\mathsf{d}$ determined at test time. They are combined as $\transParams=\transParams^\mathsf{s} \,+\, \bw \transParams^\mathsf{d}$, with  $\bw'' \in \RR^D$ a vector of learned weights. The dynamic weights can therefore be seen as an adjustment made to the static ones depending on the input $\bh$.

A set of candidate dynamic weights are maintained in an associative memory $\M$. This memory is a large set (as large as the support set, see \sect\ref{sec:support}) of key/value pairs $\M=\{(\tilde{\bh_i},\tilde{\transParams}^\mathsf{d}_i)\}_{i\in 1\ldots|\setfont{S}|}$. The interpretation for $\tilde{\transParams}^\mathsf{d}_i$ is of dynamic weights suited to an input similar to $\tilde{\bh_i}$. Therefore, at test time, we retrieve appropriate dynamic weights $\transParams^\mathsf{d}$ by soft key matching:
\begin{align}
  \label{eq:memory}
  \transParams^\mathsf{d} ~~=~~ \sum_i \tilde{\transParams}^\mathsf{d}_i \; \softmax_i\big( d_\mathsf{cos}( \bh, \tilde{\bh_i} ) \big)
\end{align}
where $d_\mathsf{cos}(\cdot,\cdot)$ is the cosine similarity function. We therefore retrieve a weighted sum, in which the similarity of $\bh$ with the memory keys $\tilde{\bh_i}$ serves to weight the memory values $\tilde{\transParams}^\mathsf{d}_i$. In practice and for computational reasons, the softmax function cuts off after the top $k$ largest values, with $k$ in the order of a thousand elements (see \sect\ref{sec:experiments}). We detail in \sect\ref{sec:support} how the memory is filled by processing the support set. Note that the above formulation can be made equivalent to the original model in~\cite{teney2017challenge} by using only static weights ($\transParams=\transParams^\mathsf{s}$). This serves as a baseline in our experiments (see \sect\ref{sec:experiments}).

\subsubsection{Mapping to Candidate Answers $g_\ProtoSet(\cdot)$}

The function $g_\ProtoSet(\bh)$ maps the output of the non-linear transformation to a vector of scores $s\in [0,1]^A$ over the set of candidate answers. It is traditionally implemented as a simple affine or linear transformation (\ie a matrix multiplication). We generalize the definition of $g_\ProtoSet(\bh)$ by interpreting it as a similarity measure between its input $\bh$ and prototypes $\ProtoSet=\{\proto_i^a\}_{i,a}$ representing the possible answers. In traditional models, each prototype corresponds to one row of the weight matrix. Our general formulation allows one or several prototypes per possible answer $a$ as $\{\proto_i^a\}_{i=1}^{N^a}$. Intuitively, the prototypes represent the typical expected feature vector when $a$ is a correct answer. The score for $a$ is therefore obtained as the similarity between the provided $\bh'$ and the corresponding prototypes of $a$. When multiple prototypes are available, the similarities are averaged. Concretely, we define
\begin{align}
  g^a_\ProtoSet(\bh') ~~=~~ \sigma\Big( \, \frac{1}{N^a} \sum^{N^a}_{i=1} d(\bh', \proto_i^a) \, + b'' \, \Big)
 \label{eq:output}
\end{align}
where $d(\cdot,\cdot)$ is a similarity measure, $\sigma$ is a sigmoid (logistic) activation function to map the similarities to $[0,1]$, and $b''$ is a learned bias term. The traditional models that use a matrix multiplication~\cite{jabri2016revisiting,teney2017challenge,teney2016zsvqa} correspond to $g_\ProtoSet(\cdot)$ that uses a dot product as the similarity function. In comparison, our definition generalizes to multiple prototypes per answer and to different similarity measures. Our experiments evaluate the dot product and the weighted L-p norm of vector differences:
\begin{align}
   d_\mathsf{dot}(\bh,\transParams) & ~~=~~ \bh^\intercal \, \transParams \\
   d_\mathsf{L1}(\bh,\transParams) & ~~=~~ \bw'''^\intercal \, \left| \bh - \transParams \right| \\
   d_\mathsf{L2}(\bh,\transParams) & ~~=~~ \bw'''^\intercal \, ( \bh - \transParams )^2 \label{eq:similarity}
\end{align}
where $\bw''' \in \RR^D$ is a vector of learned weights applied coordinate-wise.

Our model uses two sets of prototypes, the static $\ProtoSet^\mathsf{s}$ and the dynamic $\ProtoSet^\mathsf{d}$. The static ones are learned during training as traditional weights by backpropagation and gradient descent, and held fixed at test time. The dynamic ones are determined at test time by processing the provided support set (see \sect\ref{sec:support}). Thereafter, all prototypes $\ProtoSet=\ProtoSet^\mathsf{s} \,\cup\, \ProtoSet^\mathsf{d}$ are used indistinctively. Note that our formulation of $g_\ProtoSet(\cdot)$ can be made equivalent to the original model of~\cite{teney2017challenge} by using only static prototypes ($\ProtoSet=\ProtoSet^\mathsf{d}$) and the dot-product similarity measure $d_\mathsf{dot}(\cdot,\cdot)$. This will serve as a baseline in our experiments (\sect\ref{sec:experiments}).

Finally, the output of the network is attached to a cross-entropy loss $\loss(\bs,\hat{\bs})$ between the predicted and ground truth for training the model end-to-end~\cite{teney2017challenge}.

\subsection{Processing of Support Set}
\label{sec:support}

Both functions $f_\transParams(\cdot)$ and $g_\ProtoSet(\cdot)$ defined above use dynamic parameters that are dependent on the support set. Our model processes the entire support set in a forward and backward pass through the network as described below. This step is to be carried out once at test time, prior to making predictions on any instance of the test set. At training time, it is repeated before every epoch to account for the evolving static parameters of the network as training progresses (see Algorithm~\ref{alg}).

We pass all elements of the support set $\setfont{S}$ through the network in mini-batches for both a forward and backward pass. The evaluation of $f_\transParams(\cdot)$ and $g_\ProtoSet(\cdot)$ use \emph{only static weights and prototypes}, \ie $\transParams=\transParams^\mathsf{s}$ and $\proto=\proto^\mathsf{s}$. To fill the memory $\M$, we collect, for every element of the support set, its feature vector $\bh$ and the gradient $\nabla_{\transParams^\mathsf{s}} \loss$ of the final loss relative to the static weights $\transParams$. This effectively captures the adjustments that would be made by a gradient descent algorithm to those weights for that particular example. The pair $(\bh,\nabla_{\transParams^\mathsf{s}} \loss)$ is added to the memory $\M$, which thus holds $|\setfont{S}|$ elements at the end of the process.

To determine the set of dynamic prototypes $\proto^\mathsf{d}$, we collect the feature vectors $\bh'=f_\transParams(\bh)$ over all instances of the support set. We then compute their average over instances having the same correct answer. Concretely, the dynamic prototype for answer $a$ is obtained as $\proto^a = \frac{1}{N^a} \sum_{i:\hat{s}_i^a=1}^{N^a} \bh'_i$.

During training, we must balance the need for data to train the static parameters of the network, and the need for an ``example'' support set, such that the network can learn to use novel data. If the network is provided with a fixed, constant support set, it will overfit to that input and be unable to make use of novel examples at test time. Our training procedure uses all available data as the training set $\setfont{T}$, and we form a different support set $\setfont{S}$ at each training epoch as a random subset of $\setfont{T}$. The procedure is summarized in Algorithm~\ref{alg}. Note that in practice, it is parallelized to process instances in mini-batches rather than individually.

\begin{algorithm}[t]
  \KwIn{}
  Support set ~$\mathscr{S}\,=\,\{ (\mathsf{Q}_i,\mathsf{I}_i,\hat{\bs}_i) \}_i$\\
  Instance to evaluate $(\mathsf{Q},\mathsf{I},\boldsymbol{\hat{s}})$ from training or test set\\
  \KwOut{Predicted scores $\bs$ over candidate answers}
  $\Phi^\mathsf{d} \leftarrow \emptyset$~~//~\textit{Initialize dynamic prototypes}\\
  $\M \leftarrow \emptyset$~~//~\textit{Initialize memory of dynamic weights}\\
  \For{each element $i$ in support set $\mathscr{S}$}{
    \If{training, with probability $p$}{
      \textbf{continue}~~//~\textit{Drop random support elements}\\
    }
    Forward and backward propagation of $(\mathsf{Q}_i,\mathsf{I}_i,\hat{\bs}_i)$\\
    ~~~~using only static weights $\theta^\mathsf{s}$ and prototypes $\Phi^\mathsf{s}$\\
    Collect $\bh_i$\\
    ~~~~~~~~~~~~~$\bh'_i = f_\theta(\bh_i)$\\
    ~~~~~~~~~~~~~$\bs_i = g_\Phi( \bh'_i )$\\
    ~~~~~~~~~~~~~$\nabla_{\theta^\mathsf{s}} \mathscr{L}(\bs_i,\hat{\bs}_i)$\\
    //~\textit{Store (key, value) in memory of dynamic weights}\\
    $\M \,\leftarrow\, \M ~~\;\cup\,~~ (\bh, \;\nabla_{\theta^\mathsf{s}} \mathscr{L}(\bs_i,\hat{\bs}_i))$
  }
  $\phi^a ~=~ \frac{1}{N^a} \sum_{i:\hat{s}_i^a=1}^{N^a} \bh'_i~~~\forall a$~~//~\textit{One average per answer}\\
  $\Phi^\mathsf{d} ~=~ \cup_a \phi^a$~~//~\textit{Store dynamic prototypes}\\
  Forward prop. of ($\mathsf{Q}$, $\mathsf{I}$) with static and dynamic param.\\
  \uIf{test time}{
    Return predicted scores $\bs = g_\Phi\big( f_\theta(\bh) \big)$  
  }\uElseIf{training time}{
    Backpropagation and gradient descent\\
    Update static parameters $\theta^\mathsf{s}$, $\phi^\mathsf{s}$, $b''$, $w''$, $w'''$, and those of the question and image embeddings
  }\textbf{end}
  \caption{Evaluation of one test or training instance\label{alg}.}
\end{algorithm}
\vvspace{-6pt}

\section{Experiments}
\label{sec:experiments}

We perform a series of experiments to evaluate (1)~how effectively the proposed model and its different components can use the support set, (2)~how useful novel support instances are for VQA, (3)~whether the model learns different aspects of a dataset from classical VQA methods trained in the classical setting.


\paragraph{Datasets} The \emph{VQA v2} dataset~\cite{goyal2016balanced} serves as the principal current benchmark for VQA. The heavy class imbalance among answers makes it very difficult to draw meaningful conclusions or perform a qualitative evaluation, however. We additionally propose a series of experiments on a subset referred to as \emph{VQA-Numbers}. It includes all questions marked in \emph{VQA v2} as a ``number'' question, which are further cleaned up to remove answers appearing less than 1,000 times in the training set, and to remove questions that do not have an unambiguous answer (we keep only those with ground truth scores containing a single element equal to $1.0$). Questions from the original validation set of \emph{VQA v2} are used for evaluation, and the original training set (45,965 questions after clean up) is used for training, support, and validation. The precise data splits will be available publicly. Most importantly, the resulting set of candidate answers corresponds to the seven numbers from 0 to 6.

\paragraph{Metrics} The aggregate metric used for evaluation on \emph{VQA v2} is the accuracy defined as $\frac{1}{|\mathscr{E}|} \sum_i \hat{s}_i^{a^\star_i}$ with ground truth scores $\hat{s}_i$ and ${a^\star_i}$ the answer of highest predicted score, $\argmax_a s_i^a$. We also define the recall of an answer $a$ as $\sum_i s_i^{a^\star} ~~/~~ \sum_i \hat{s}_i^a$. We look at the recall averaged (uniformly) over all possible answers to better reflect performance across a variety of answers, rather than on the most common ones.

\paragraph{Implementation}

Our implementation is based on the code provided by the authors of~\cite{teney2017challenge}. Details non-specific to our contributions can be found there. We initialize all parameters, in particular static weights and static prototypes as if they were those of a linear layer in a traditional architecture, following Glorot and Bengio~\cite{glorot2010understanding}. During training, the support set is subsampled (\sect\ref{sec:support}) to yield a set of 1,000 elements. We use, per answer, one or two static prototypes, and zero or one dynamic prototype (as noted in the experiments). All experiments use an embedding dimension $D$=128 and a mini-batches of $256$ instances. Experiments with \emph{VQA v2} use a set of candidate answers capped to a minimum number of training occurrences of $16$, giving 1,960 possible answers~\cite{teney2017challenge}. Past works have shown that small differences in implementation can have noticeable impact on performance. Therefore, to ensure fair comparisons, we repeated all evaluations of the baseline~\cite{teney2017challenge} with our code and preprocessing. Results are therefore not directly comparable with those reported in~\cite{teney2017challenge}. In particular, we do not use the \emph{Visual Genome} dataset~\cite{krishnavisualgenome} for training.

\subsection{VQA-Numbers}
\label{sec:vqaNumbers}

\paragraph{Ablative evaluation}
\label{sec:ablations}
We first evaluate the components of the proposed model in comparison to the state-of-the-art of~\cite{teney2017challenge} which serves as a baseline, being equivalent to our model with 1 static prototype per answer, the dot product similarity, and no dynamic parameters. We train and evaluate on all 7 answers.  To provide the baseline with a fair chance\footnote{The \emph{VQA-Numbers} data is still heavily imbalanced, ``1'' and ``2'' making up almost $60\%$ of correct answers in equal parts.}, we train all models with standard supersampling~\cite{buda2017imbalance,guo2017review}, \ie selecting training examples with equal probability with respect to their correct answer. In these experiments, the support set is equal to the training set.

\begin{table}[t]
\scriptsize
\renewcommand{\arraystretch}{1.36}
\centering
\begin{tabularx}{\linewidth}{Xc}
\Xhline{1\arrayrulewidth}
\multicolumn{2}{r}{Average answer recall} \\
\Xhline{1\arrayrulewidth}
\hspace{-.5em}(1a)~~~Chance  & \multicolumn{1}{l}{14.28~~~~~~} \\
\hspace{-.5em}(1b)~~~State-of-the-art model \cite{teney2017challenge} & \multicolumn{1}{l}{29.72~~~~~~} \\
\multicolumn{2}{l}{\hspace{-.5em}Equivalent to 1 static prototype per answer, dot prod. similarity, no dynamic param.} \\
\hline
\hspace{-.5em}(2b)~~~1 Static prot./ans., L1 similarity        & \multicolumn{1}{l}{29.97~~~~~~} \\
\hspace{-.5em}(2c)~~~1 Static prot./ans., L2 similarity        & \multicolumn{1}{l}{27.80~~~~~~} \\
\hspace{-.5em}(2d)~~~2 Static prot./ans., dot prod. similarity & \multicolumn{1}{l}{30.28~~~~~~} \\
\hspace{-.5em}(2e)~~~2 Static prot./ans., L1 similarity        & \multicolumn{1}{l}{28.34~~~~~~} \\
\hspace{-.5em}(2f)~~~2 Static prot./ans., L2 similarity        & \multicolumn{1}{l}{31.48~~~~~~} \\
\hspace{-.5em}(3a)~~~Dynamic Weights (+2f)                     & \multicolumn{1}{l}{31.81~~~~~~} \\
\hspace{-.5em}(3b)~~~\textbf{Proposed: dynamic weights and prototypes} (+2f)    & \multicolumn{1}{l}{\textbf{32.32}~~~~~~} \\
\Xhline{1\arrayrulewidth}
\end{tabularx}
\normalsize
\vspace{3pt}
\normalsize
\caption{Ablative evaluation on \emph{VQA-Numbers}, trained and evaluated on all answers. See discussion in \sect\ref{sec:ablations}.}
\label{tableAblations}
\vspace{-3pt}
\end{table}

As reported in Table~\ref{tableAblations}, the proposed dynamic weights improve over the baseline, and the dynamic prototypes bring an additional improvement. We compare different choices for the similarity function. Interestingly, swapping the dot product in the baseline for an L2 distance has a negative impact. When using two static prototypes however, the L2 distances proves superior to the L1 or to the dot product. This is consistent with~\cite{snell2017proto} where a prototypes network also performed best with an L2 distance.

\begin{figure}[t]
  \centering
  \includegraphics[width=0.90\linewidth]{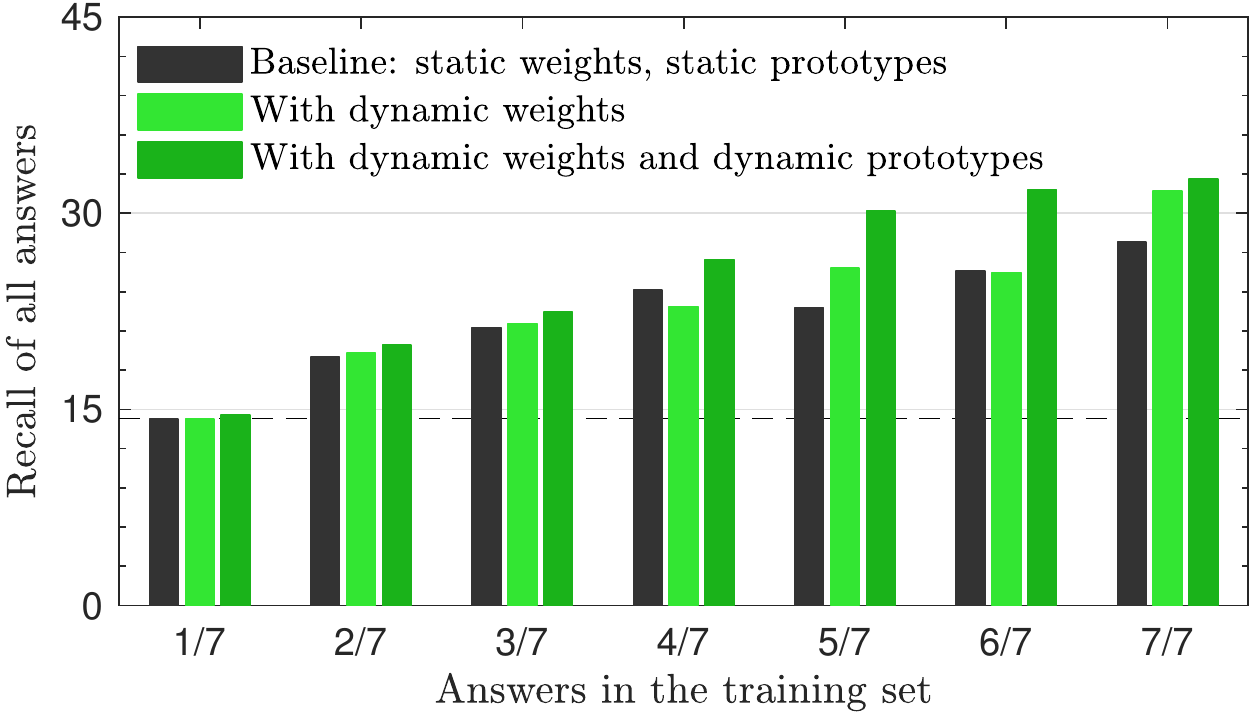}\vspace{-10pt}
  \includegraphics[width=0.90\linewidth]{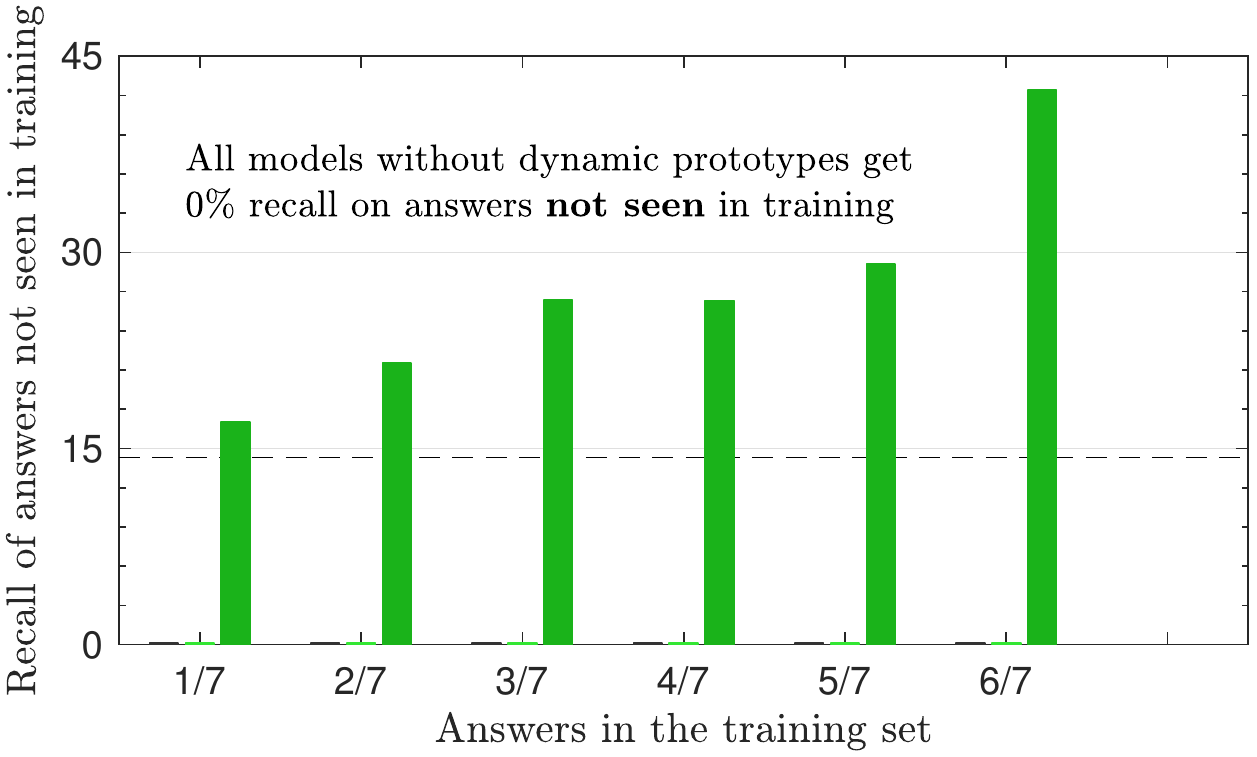}
  \vvspace{0pt}
  \caption{Performance of the proposed model and ablations on \emph{VQA-Numbers} with training data for a subset of the 7 answers. (Top)~Performance on all answers. (Bottom)~Performance on answers not seen in training. Only the model with dynamic prototypes makes this setting possible. Remarkably, a model trained on two answers (2/7) maintains a capacity to learn about all others. Chance baseline shown as horizontal dashes.}
  \label{fig:vqaNumbers}
  \vvspace{-4pt}
\end{figure}

\paragraph{Additional Support Set and Novel answers}
We now evaluate the ability of the model to exploit support data never seen until test time (see \fig\ref{fig:vqaNumbers}). We train the same models designed for 7 candidate answers, but only provide them with training data for a subset of them. The proposed model is additionally provided with a complete support set, covering all 7 answers. Each reported result is averaged over 10 runs. The set of $k$ answers excluded from training is randomized across runs but identical to all models for a given $k$.

The proposed model proves superior than the baseline and all other ablations (\fig\ref{fig:vqaNumbers}, top). The dynamic prototypes are particularly beneficial. With very little training data, the use of dynamic weights is less effective and sometimes even detrimental. We hypothesize that the model may then suffer from overfitting due to the additional learned parameters. When evaluated on novel answers (not seen during training and only present in the test-time support set), the dynamic prototypes provide a remarkable ability to learn those from the support set alone (\fig\ref{fig:vqaNumbers}, bottom). Their efficacy is particularly strong when only a single novel answer has to be learned. Remarkably, a model trained on only two answers maintains some capacity to learn about all others (average recall of $17.05\%$, versus the chance baseline of $14.28\%$). Note that we cannot claim the ability of the model to count to those novel numbers, but at the very least it is able to associate those answers with particular images/questions (possibly utilizing question-conditioned biases).

\subsection{VQA v2}
\label{sec:vqav2}

\begin{figure}[t]
  \centering
  \includegraphics[width=0.90\linewidth]{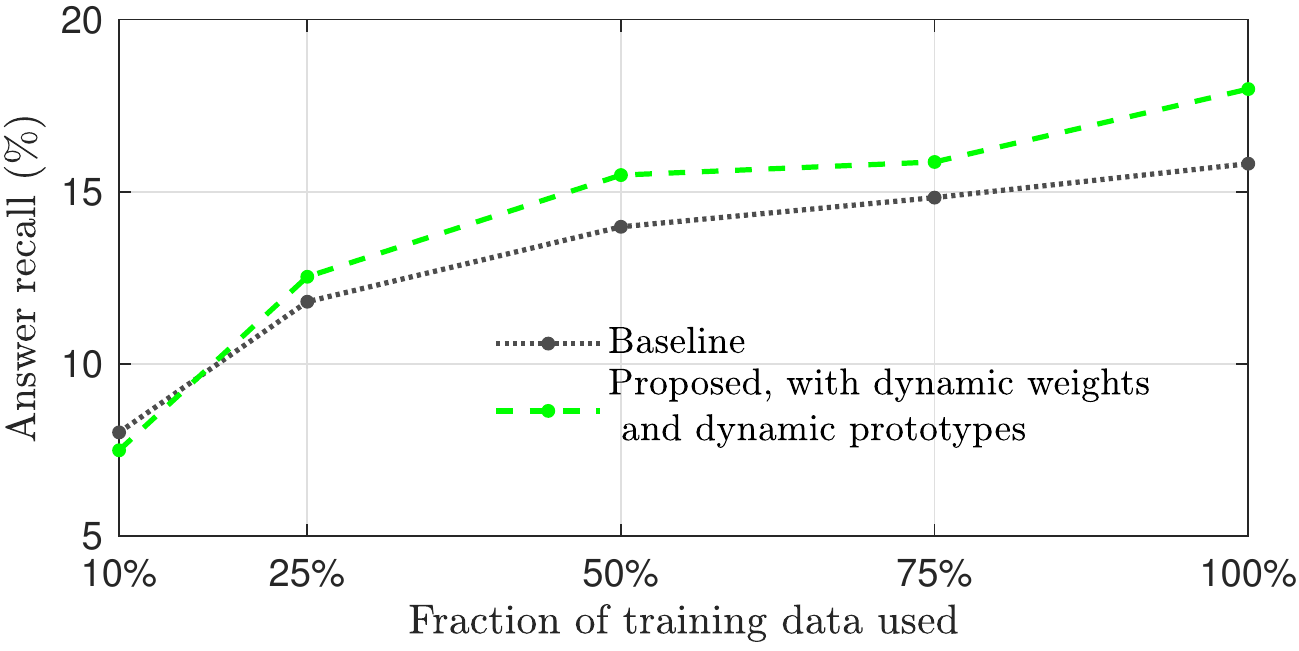}
  \caption{Performance using varying amounts of training data on \emph{VQA v2}. See \sect\ref{sec:vqav2}.}
  \label{fig:trData}
  \vvspace{-4pt}
\end{figure}

We performed experiments on the complete \emph{VQA v2} dataset. We report results of different ablations, trained with 50\% or 100\% of the official training set, evaluated on the validation set as in~\cite{teney2017challenge}. The proposed model uses the remaining of the official training set as additional support data at test time. The complexity and varying quality of this dataset do not lead to clear-cut conclusions from the standard accuracy metric (see Table~\ref{tableVqaFull}). The answer recall leads to more consistent observations that align with those made on \emph{VQA-Numbers}. Both dynamic weights and dynamic parameters provide a consistent advantage (\fig\ref{fig:trData}). Each technique is beneficial in isolation, but their combination performs generally best. Individually, the dynamic prototypes appear more impactful than the dynamic weights. Note that our experiments on \emph{VQA v2} aimed at quantifying the effect of the contributions in the meta learning setting, and we did not seek to maximize absolute performance in the traditional benchmark setting.


\begin{table}[t]
\scriptsize
\renewcommand{\tabcolsep}{0.05em}
\renewcommand{\arraystretch}{1.36}
\centering
\begin{tabularx}{\linewidth}{Xcc}
\Xhline{1\arrayrulewidth}
                                          &   \multicolumn{2}{c}{Question accuracy ~~/~ Answer recall} \\ \cline{2-3} 
                                          &   Trained on 50\% & Trained on 100\% \\
\Xhline{1\arrayrulewidth}

Baseline \cite{teney2017challenge} &  \textbf{57.6} / 14.0 & 59.8 / 15.8 \\
\hline
Proposed model\\
\textbf{With} dynamic weights, no dynamic prototypes &  \textbf{57.6} / 14.1 & \textbf{60.0} / 16.3 \\
No dynamic weights, \textbf{with} dynamic prototypes &  \textbf{57.6} / 15.2 & 59.7 / \textbf{18.0} \\
~~~Same, no static prototypes, only dyn. ones &   57.2 / ~3.6 & 58.6 / ~4.29 \\
\textbf{With} dyn. weights and dyn. prototypes &  57.5 / \textbf{15.5} & 59.9 / \textbf{18.0} \\

\Xhline{1\arrayrulewidth}
\end{tabularx}
\normalsize
\normalsize
\caption{Evaluation on \emph{VQA v2}. The proposed method exhibits qualitatively different strengths than the classical approach of the baseline~\cite{teney2017challenge}, producing a generally higher recall (averaged over possible answers) and lower accuracy (averaged over questions).}
\label{tableVqaFull}
\end{table}

\begin{figure}[t]
  \centering
  \includegraphics[width=0.90\linewidth]{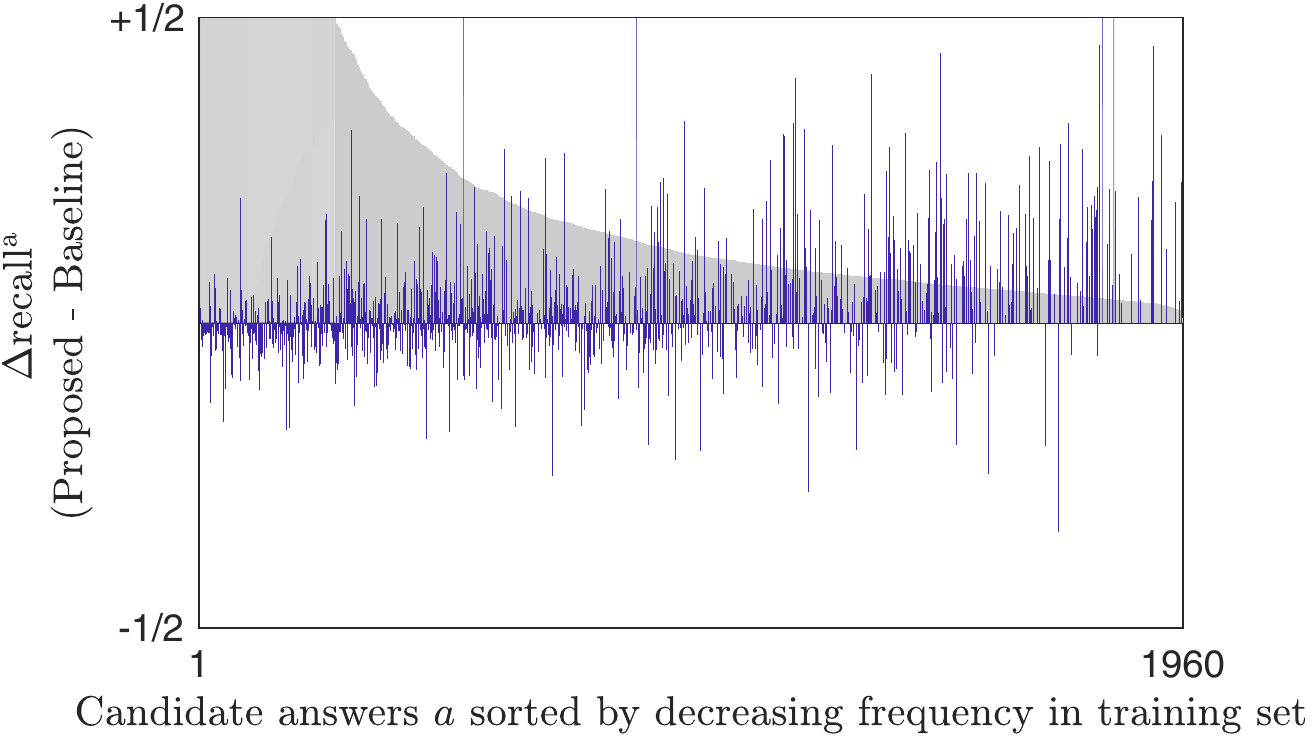}
  \vvspace{0pt}
  \caption{Difference in answer recall between the proposed model (Table~\ref{tableVqaFull}, last row, last column) and the baseline (Table~\ref{tableVqaFull}, first row, last column) on \emph{VQA v2}. Each blue bar corresponds to one of the candidate answers, sorted decreasing number of occurrences in the training set (gray background, units not displayed). The two models show qualitatively different behaviour: the baseline is effective with frequent answers, but the proposed model fares better (mostly positive values) in the long tail of rare answers.}
  \label{fig:answerRecall}
  \vvspace{-1pt}
\end{figure}


To obtain a better insight into the predictions of the model, we examine the individual recall of possible answers. We compare the values with those obtained by the baseline. The difference (\fig\ref{fig:answerRecall}) indicates which of the two models provides the best predictions for every answer. We observe a qualitatively different behaviour between the models. While the baseline is most effective with frequent answers, the proposed model fares better (mostly positive values) in the long tail of rare answers. This corroborates previous discussions on dataset biases~\cite{goyal2016balanced,jabri2016revisiting,zhang2015balanced} which classical models are prone to overfit to. The proposed model is inherently more robust to such behaviour.

\section{Conclusions and Future Work}

We have devised a new approach to VQA through framing it as a meta learning task. This approach enables us to provide the model with supervised data at test time, thereby allowing the model to adapt or improve as more data is made available. We believe this view could lead to the development of scalable VQA systems better suited to practical applications. We proposed a deep learning model that takes advantage of the meta learning scenario, and demonstrated a range of benefits: improved recall of rare answers, better sample efficiency, and a unique capability of to learn to produce novel answers, \ie those never seen during training, and learned only from support instances.

The learning-to-learn approach we propose here enables a far greater separation of the questions answering method from the information used in the process than has previously been possible.  Our contention is that this separation is essential if vision-and-language methods are to move beyond benchmarks to tackle real problems, because embedding all of the information a method needs to answer real questions in the model weights is impractical.

Even though the proposed model is able to use novel support data, the experiments showed room for improvement, since a model trained initially from the same amount of data still shows superior performance. Practical considerations should also be addressed to apply this model to a larger scale, in particular for handling the memory of dynamic weights that currently grows linearly with the support set. Clustering schemes could be envisioned to reduce its size~\cite{snell2017proto} and hashing methods~\cite{andoni2006near,kaiser2017rare} could improve the efficiency of the content-based retrieval.

Generally, the handling of additional data at test time opens the door to VQA systems that interact with other sources of information. While the proposed model was demonstrated with a support set of questions/answers, the principles extend to any type of data obtained at test time \eg from knowledge bases or web searches. This would drastically enhance the scalability of VQA systems.


{\small\bibliographystyle{ieee}\bibliography{Bibliography}}
\clearpage

\appendix
\section*{Supplementary material}
\vspace{12pt}

\label{techDetails}

\section{Factorized Non-Linear Transformation}

We defined in \eq\ref{eq:nonLinearLayer} our non-linear transformations using weights (static or dynamic) $\transParams$ containing all the parameters of a gated tanh layer. Although this is sound in principle, it is computationally costly to handle a memory containing dynamic weights of such a large dimensionality ($2D^2+2D$). To alleviate this, we follow~\cite{BertinettoHVTV16} and factorize the parameters of the gated tanh layer, rewriting \eq\ref{eq:nonLinearLayer} as follows:
\begin{align}
  \label{eq:nonLinearLayer2}
  f_\transParams(\bh) ~~=~~ \sigma( \bw_a W_b \bh + \bb ) ~\circ~ \tanh\,(\bw'_a W_b' \bh + \bb' )
\end{align}
with vectors $\bw_a, \bw'_a \in \RR^D$ and matrices $W_b, W'_b\in \RR^{D \times D}$. The matrices $W_b$ and $W'_b$ are learned like traditional weights, and only $\bw_a$ and $\bw'_a$ are those incorporated into the vector of weights $\transParams$ (static or dynamic). This reduces the dimensionality of $\transParams$ from $(2D^2+2D)$ to $4D$ (accounting for $\bw_a$, $\bw_b$, $\bb$, and $\bb'$).

\vspace{20pt}

\section{Bias in the Output Mapping}
In \eq\ref{eq:output}, the output mapping to answer scores uses a \emph{scalar} bias term. This is different to the vector bias in a classical model that use a linear (affine) layer to implement $g_\ProtoSet(\cdot)$. A vector contains a value for each output class (\ie each candidate answer), whereas our formulation uses a single value shared among all of them. Our formulation helps avoid the model incorporating biases towards frequent training answers, as discussed in \sect\ref{sec:intro} and \ref{sec:related}. This feature is essential to enable the capability of our model to produce novel answers (unseen during training) only demonstrated by instances in the support set. A vector of answer-specific biases would prevent this capability, as the bias for the novel answers can not be learned from training data.

Completely removing the bias term is another option. At test time, it is without effect compared to a scalar bias, since it is only followed by a (monotonic) sigmoid. Removing the bias term however renders the training by gradient descent numerically unstable, because the chosen similarity function can map to saturating regions of the domain of the sigmoid.



\newpage
\section{VQA-Numbers Dataset}

\noindent
We provide below statistics of the \emph{VQA-Numbers} dataset.

\begin{table}[h!]
\scriptsize
\renewcommand{\tabcolsep}{0.95em}
\renewcommand{\arraystretch}{1.36}
\centering
\begin{tabularx}{\linewidth}{ccccccccc}
\Xhline{1\arrayrulewidth}
\multicolumn{7}{c}{Correct answer} & Row \vspace{-2pt}\\ 
0 & 1 & 2 & 3 & 4 & 5 & 6 & sum\\
\Xhline{1\arrayrulewidth}
\multicolumn{7}{l}{Training/validation/support set} \\
2,529 & 8,193 & 7,030 & 2,485 & 1,520 &  579 &  602 & 22,938 \\
11.0\% & 35.7\% & 30.7\% & 10.8\% &  6.6\% &  2.6\% &  2.6\% & 100\% \\
\Xhline{1\arrayrulewidth}
\multicolumn{7}{l}{Test set} \\
858 & 2,804 & 2,434 &  843 &  495 &  173 &  205 & 7,812 \\
11.0\% & 35.9\% & 31.2\% & 10.8\% &  6.3\% &  2.2\% &  2.6\% & 100\% \\
\Xhline{1\arrayrulewidth}
\end{tabularx}
\normalsize
\vspace{9pt}
\normalsize
\caption{Number and ratio of questions per type of correct answer in the \emph{VQA-Numbers} dataset.}
\label{tableVqaNumbersStats}
\end{table}

\section{VQA-Numbers Experiments}

Our experiments on \emph{VQA-Numbers} (\sect\ref{sec:vqaNumbers}) use supersampling during training to ensure that none of the compared models can be influenced by dataset biases (\ie class imbalance). The supersampling is performed at the epoch level, not at the mini-batch level. Concretely, the training instances of all answers (classes) except the most frequent one are repeated at random such that there are as many of them as instances with the most frequent one. The elements within a mini-batch are selected at random. Each training epoch will thus go through every training instance at least once. We did not try to constrain the sampling within mini-batches.

Note finally that the supersampling strategy is practical on VQA-Numbers thanks to the small number of classes and only mild imbalance. It would not be suitable to \emph{VQA v2} for the opposite reasons.

\begin{figure}[h!]
  \centering
  \includegraphics[width=0.90\linewidth]{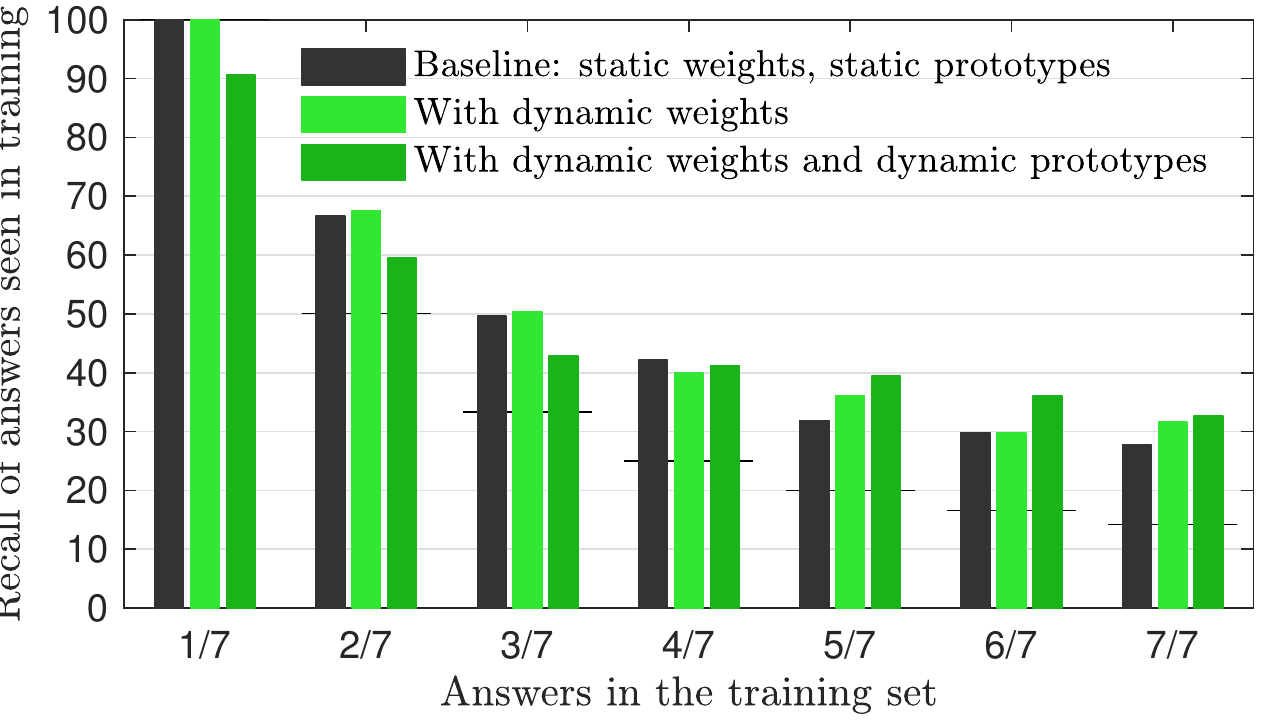}
  \caption{Additional results of the experiments of \fig\ref{fig:vqaNumbers}. Performance is reported here only on answers seen in the training set. See text for discussion.}
  \label{fig:vqaNumbers2}
  \vspace{-4pt}
\end{figure}

We provide in \fig~\ref{fig:vqaNumbers2} additional results of the experiments of \sect\ref{sec:vqaNumbers}. We report the performance of the same models, now evaluated only on answers present in the training set. The number of those answers is varied from 1 to 7. The chance performance (gray dashes) diminishes as the number of possible answers gets larger. As expected, the baseline model performs at 100\% recall in the trivial case of 1 possible answer. The proposed model however receives a support set containing examples of all other answers (\ie all 7 of them). This explains the non-perfect result in the trivial case. As the number of possible answers is increased, the proposed models (with dynamic weights and dynamic prototypes) shows a growing advantage and finally surpass the baseline by a clear margin on the most interesting cases of 5, 6, and 7 answers.

\end{document}